\documentclass[letterpaper, 10 pt, conference]{ieeeconf}  

\IEEEoverridecommandlockouts                              
\usepackage[utf8]{inputenc}
\usepackage[T1]{fontenc}

\usepackage[table]{xcolor}
\definecolor{rblue}{rgb}{0,0.5,1}
\definecolor{awesome}{rgb}{1.0, 0.13, 0.32}
\definecolor{hollywoodcerise}{rgb}{0.96, 0.0, 0.63}
\definecolor{lasallegreen}{rgb}{0.03, 0.47, 0.19}
\definecolor{hanpurple}{rgb}{0.32, 0.09, 0.98}
\definecolor{green(pigment)}{rgb}{0.0, 0.65, 0.31}

\makeatletter
\let\NAT@parse\undefined
\makeatother
\usepackage[pagebackref=false, breaklinks=true, colorlinks, bookmarks=false]{
        hyperref
}
\hypersetup{
        colorlinks=true,
        linkcolor={red},
        citecolor={hanpurple},
        urlcolor={magenta}
}

\usepackage{caption}
\usepackage{graphicx}
\usepackage{amsmath}      
\usepackage{amssymb}      
\usepackage{subcaption}   
\usepackage{booktabs}     
\usepackage{wrapfig}
\usepackage{float}
\let\labelindent\relax
\usepackage{enumitem}

\usepackage{multirow}
\usepackage{bbm}          
\usepackage{cleveref}     
\usepackage{capt-of}
\usepackage{makecell}
\usepackage{dblfloatfix}

\crefname{table}{Tab.\@}{Tabs.\@}
\crefname{figure}{Fig.\@}{Figs.\@}
\crefname{section}{Sec.\@}{Secs.\@}

\creflabelformat{table}{#2#1#3}
\creflabelformat{figure}{#2#1#3}
\creflabelformat{section}{#2#1#3}

\renewcommand{\thetable}{\arabic{table}}
\renewcommand{\thefigure}{\arabic{figure}}

\title{\LARGE \bf
CoRef-GS: Cooperative Referring Gaussian Splatting for Multi-Agent Scene Understanding
}

\author{Zhikun Zhou$^{1}$, Kunyu Peng$^{2,3,\dag}$, Runyi Yang$^{3}$, Junhao Cai$^{1}$, Di Wen$^{2}$, Ruiping Liu$^{2}$,\\Danda Pani Paudel$^{3}$, Yi Zhou$^{1}$, Luc Van Gool$^{3}$, and Kailun Yang$^{1,\dag}$
\thanks{This work was supported in part by the National Natural Science Foundation of China (Grant No. 62473139, No. 62673195, and No. 62388101), in part by the Hunan Provincial Research and Development Project (Grant No. 2025QK3019), in part by the State Key Laboratory of Autonomous Intelligent Unmanned Systems (the opening project number ZZKF2025-2-10), and in part by the Deutsche Forschungsgemeinschaft (DFG, German Research Foundation) – SFB-1574 – 471687386.}
\thanks{$^{1}$The authors are with the School of Artificial Intelligence and Robotics and the National Engineering Research Center of Robot Visual Perception and Control Technology, Hunan University, China (email: kailun.yang@hnu.edu.cn).}%
\thanks{$^{2}$The authors are with the Institute for Anthropomatics and Robotics, Karlsruhe Institute of Technology, Germany (email: kunyu.peng@kit.edu).}
\thanks{$^{3}$The authors are with INSAIT, Sofia University ``St. Kliment Ohridski'', Bulgaria.}
\thanks{$^{\dag}$Corresponding authors: Kailun Yang and Kunyu Peng.}
}

\begin{document}
\raggedbottom

\let\oldtwocolumn\twocolumn
\renewcommand{\twocolumn}[1][]{%
    \oldtwocolumn[{#1%
        \centering
        \includegraphics[width=0.99\textwidth]{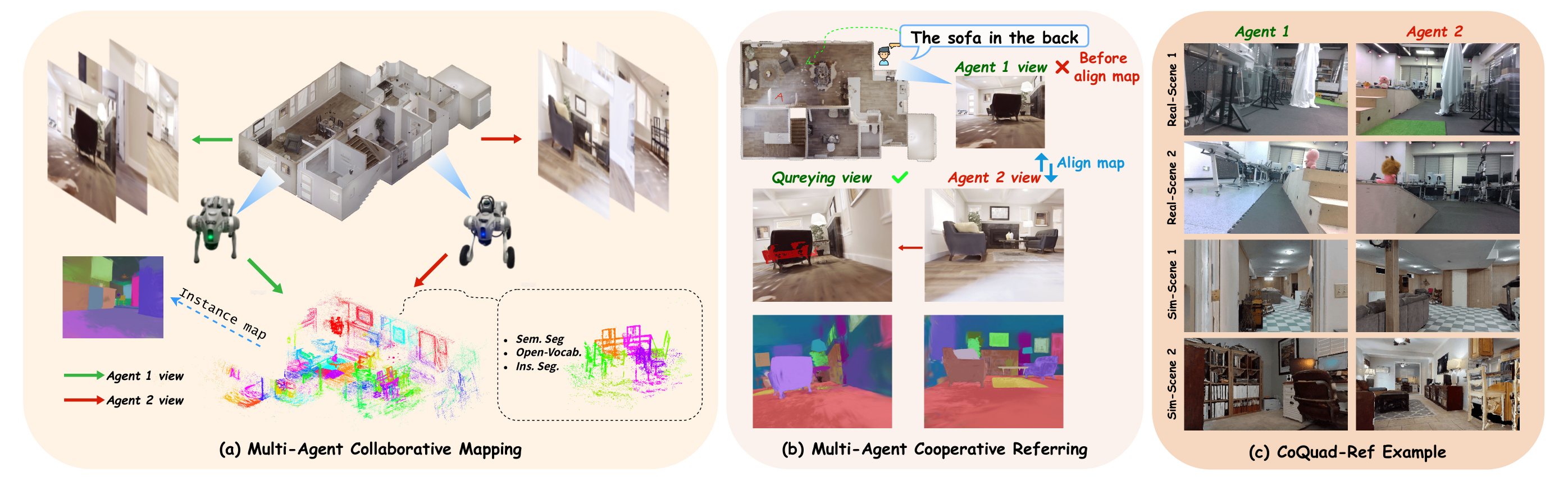}
        \vskip-2ex
        \captionof{figure}{\textbf{Overview of CoRef-GS.}
        (a) Multiple quadruped agents independently build local open-vocabulary semantic Gaussian maps from complementary observations.
        CoRef-GS aligns the local maps with image-assisted coarse-to-fine
        $\mathrm{Sim}(3)$ registration and directly fuses them without retraining.
        (b) Given a robot-centered referring query, CoRef-GS renders the fused map from the querying view and performs relation-aware grounding, enabling segmentation of targets that are unobserved or ambiguous in a single-agent map.
        (c) Representative CoQuad-Ref examples from real-world and simulated scenes.}
        \label{fig:teaser}
    }]%
}
\maketitle
\let\twocolumn\oldtwocolumn
\thispagestyle{empty}
\pagestyle{empty}

\begin{abstract}
Referring scene understanding for embodied robots requires grounding object- and relation-centric language queries from a designated viewpoint. While a local semantic Gaussian map can support such grounding within one agent's observations, cooperative settings require this ability to remain effective after independently reconstructed maps are aligned and fused. In this setting, the referred target or its contextual landmark may come from another agent's observations, while spatial relations must still be interpreted from the querying robot's viewpoint. We formulate this problem as cooperative referring Gaussian grounding over fused maps, which requires geometric alignability, instance-level semantic comparability, and view-conditioned relation reasoning. Existing language-aware Gaussian methods mainly focus on single-map querying, whereas Gaussian registration methods optimize geometric or photometric alignment without preserving language-grounding-oriented semantic compatibility. We propose \textbf{CoRef-GS}, a cooperative referring Gaussian splatting framework. CoRef-GS constructs local open-vocabulary instance-aware Gaussian maps, then aligns partially overlapping maps with a cross-agent alignment module by geometric and semantic consistency, and grounds queries using a view-conditioned mask relation graph. We further introduce \textbf{CoQuad-Ref}, a dual-quadruped benchmark spanning both real-world and simulated indoor scenes. Experiments show that, on simulated scenes, CoRef-GS reduces the rotation error from $2.58^{\circ}$ after coarse initialization to $0.15^{\circ}$ after refinement, and improves real-world referring mIoU over ReferSplat from $52.6\%$ to $68.8\%$. The established benchmark and source code will be publicly released at \url{https://github.com/ruojiruoli17/CoRef-GS.git}.
\end{abstract}

\section{Introduction}
Language-grounded embodied scene understanding is crucial for robots operating in realistic indoor environments~\cite{anderson2018vision,krantz2020beyond,shridhar2020alfred,chen2020scanrefer}.
Beyond single-view perception, such robots need persistent 3D scene representations that accumulate multi-view observations, preserve object-level memory, and support natural-language grounding under occlusion and partial observability. 
Recent embodied systems have increasingly explored multi-robot and multi-agent collaboration, including long-horizon quadruped manipulation~\cite{wang2026odyssey}, language-guided navigation with 3D perception~\cite{liu2026omnivln}, and cooperative decision-making through dialogue, perception, or task planning~\cite{zhou2026deconav,zha2026aircopbench,liu2025coherent}. 
These efforts suggest extending language-grounded scene understanding from individual robots to cooperative robot teams. 
However, they mainly focus on task execution, communication, or policy-level coordination, leaving open how cooperative agents should maintain persistent spatial understanding for language grounding.
This raises a representation-level challenge: each robot should build a local 3D map that is language-queryable on its own, while cross-agent alignment and fusion should preserve this grounding ability in a shared representation. 
The representation must therefore be not only geometrically alignable but also semantically compatible across independently reconstructed local maps.

Since language-queryable 3D maps require the underlying scene representation to encode language-aligned semantics, language-aware Gaussian representations provide a natural starting point for building such maps.
One line of work attaches CLIP~\cite{radford2021learning-clip} aligned language features to Gaussian fields, enabling open-vocabulary querying in reconstructed scenes~\cite{qin2024langsplat,li2025langsplatv2}. 
However, such field-level language features often struggle to distinguish fine-grained instances or support spatial-relation expressions. 
Another line of work improves object-level 3D language understanding through instance-aware features, direct language embedding registration, or object-level semantic aggregation~\cite{wu2024opengaussian,jun2025dr_splat,cen2025laga}. 
These methods provide stronger object-centric semantics, but they are still mainly designed for object-level or category-level querying within a single reconstructed map.
Referring Gaussian methods~\cite{he2025refersplat} further improve fine-grained language grounding by learning scene-specific referring features. 
However, these features are optimized within independently initialized feature spaces and are not anchored to a shared semantic basis across different local maps. 
More critically, scene-specific referring features learned independently for different local maps do not share a common query space. Consequently, geometric alignment alone does not make the maps directly query-compatible: a referring model trained for one map cannot reliably interpret semantic features reconstructed in another.
Meanwhile, Gaussian registration methods ~\cite{cheng2025reggs,yuan2024photoreg,chang2024gaussreg} mainly focus on geometric or photometric consistency, without explicitly considering whether the aligned maps remain semantically compatible for downstream language grounding.

In this paper, we study cooperative referring scene understanding, where local Gaussian maps independently reconstructed by multiple agents are aligned and fused into a shared 3D representation for grounding natural-language referring expressions from the viewpoint of a designated querying agent.
We consider a cooperative multi-robot scenario in which two agents observe the same static indoor scene from different regions, resulting in partial overlap, large viewpoint changes, and complementary local observations.

As illustrated in Fig.~\ref{fig:teaser}, the aligned representation enables the querying agent to resolve referring expressions whose target object or contextual landmark is ambiguous or absent in its own local map.
This setting differs from conventional single-agent referring segmentation because the target object or its contextual landmark may be reconstructed from another robot's observations, while the final grounding decision must still be interpreted relative to the querying robot's viewpoint.
To address these challenges, we propose \textbf{CoRef-GS}, a cooperative referring scene understanding framework for embodied multi-agent systems based on Gaussian maps.
Each agent reconstructs an instance-aware local semantic Gaussian map, where Gaussian primitives are associated with CLIP-aligned semantic anchors and instance-level grouping information.
We then align independently reconstructed local maps through an image-assisted coarse-to-fine registration pipeline and merge matched local instances into a shared global representation.

At query time, rather than relying on scene-specific referring features, we render visible instance masks from the aligned Gaussian map and perform view-conditioned relation reasoning from the querying robot's perspective.
This supports agent-centric relations while retaining view-invariant object-centric ones such as \textit{above}, \textit{on}, and \textit{beneath}, preserving open-vocabulary and relation-aware grounding after fusion without jointly training scene-specific referring features across agents.

To facilitate evaluation and future research on this problem, we establish a dual-quadruped robot dataset comprising both real-world and simulated indoor scenes. In each scene, two robots observe the same environment from different regions and reconstruct partially overlapping local maps. 
Unlike common Gaussian reconstruction or registration settings~\cite{cheng2025reggs,chang2024gaussreg,yuan2024photoreg}, which often assume continuous trajectories or substantial view overlap, our dataset captures cross-agent observations with partial overlap, large viewpoint changes, and complementary scene coverage, providing a challenging testbed for map alignment, fusion, and language-grounded referring understanding. 
On this benchmark, CoRef-GS demonstrates effectiveness in both cross-agent map alignment and language-grounded referring understanding. 
CoRef-GS reduces the simulated-scene rotation error from $2.58^{\circ}$ after coarse initialization to $0.15^{\circ}$ after refinement, and improves real-world referring mIoU over ReferSplat from $52.6\%$ to $68.8\%$.
Our contributions are summarized as follows:
{
\setlength{\leftmargini}{1.3em}
\begin{itemize}
    \item We formulate \textbf{cooperative referring scene understanding} for embodied multi-agent Gaussian maps, identifying cross-agent mergeability and referability after fusion as joint requirements prior work satisfies only individually.

    \item We propose \textbf{CoRef-GS}, a cooperative referring Gaussian splatting framework that decouples semantic map construction from query-time spatial referring reasoning via image-assisted coarse-to-fine $\mathrm{Sim}(3)$ registration, geometry-semantic instance fusion, and view-conditioned relational grounding over rendered instance masks.

    \item We release \textbf{CoQuad-Ref}, a benchmark targeting cooperative referring segmentation in dual-quadruped settings with paired real-world and simulated scenes, partial cross-agent overlap, large viewpoint changes, and viewpoint-conditioned referring expressions.
\end{itemize}
}

\begin{figure*}[!t]
    \centering
    \includegraphics[width=\textwidth]{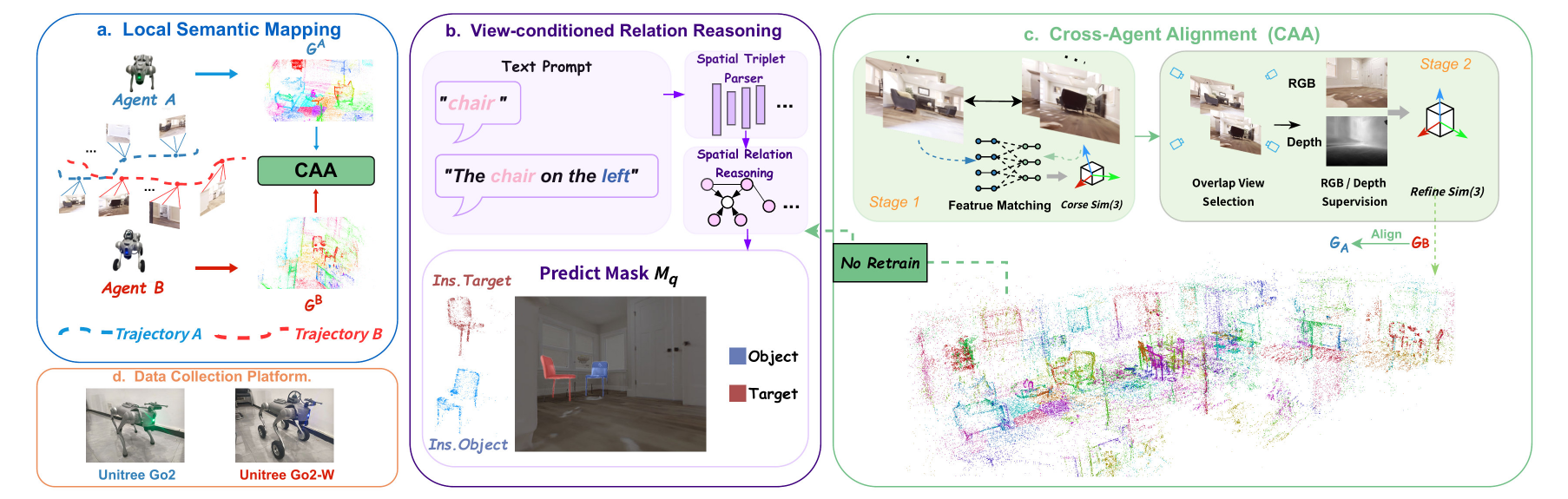}
    \captionsetup{skip=2pt}
    \caption{\textbf{The framework of CoRef-GS.}
    Agents build local instance-aware semantic Gaussian maps, which CoRef-GS aligns via image-assisted coarse-to-fine $\mathrm{Sim}(3)$ registration, fuses and reasons over for local or fused referring segmentation.}
    \vspace{-1em}
    \label{fig:Pipeline}
\end{figure*}

\section{Related Work}

\noindent\textbf{Open-Vocabulary 3D Scene Understanding.}
Open-vocabulary 3D scene understanding builds on neural and explicit scene representations, from NeRF-style radiance fields~\cite{mildenhall2021nerf} and efficient neural graphics primitives~\cite{muller2022instant} to real-time 3D Gaussian Splatting~\cite{kerbl20233d}. 
With foundation models such as CLIP~\cite{radford2021learning-clip}, DINO~\cite{caron2021dino}, SAM~\cite{kirillov2023segany}, and language-driven segmentation~\cite{li2022language-lseg}, early NeRF-based methods such as LERF~\cite{kerr2023lerf} and Open-NeRF~\cite{zhang2024opennerf} distill language or mask features into 3D fields for text-driven querying and object-level decomposition.
LangSplat~\cite{qin2024langsplat} and OpenGaussian~\cite{wu2024opengaussian} augment Gaussian primitives with language or instance-aware features. 
Follow-up works improve  practicality through efficient feature grids~\cite{ji2025fastlgs}, global-codebook feature splatting~\cite{li2025langsplatv2}, compact feature encoding~\cite{jegou2010product}, voxel-aware semantic fusion~\cite{li2025vaf_langsplat}, and Hough-voting-based localization~\cite{jiang2025votesplat}.
However, most methods remain driven by feature distillation or embedding-space matching, which limits complex referential, compositional, spatial, and relational reasoning. 
ReferSplat~\cite{he2025refersplat} and ReLaGS~\cite{xie2026relags} address this direction through 3D referential segmentation and scene-graph-based relational reasoning, but they operate within a single reconstructed scene; whether semantic or referring representations remain consistent after independently built maps are aligned remains unexplored.

\noindent\textbf{Multi-Agent 3D Gaussian Registration.}
Recent embodied multi-agent studies motivate shared scene understanding across robots, covering long-horizon quadruped exploration~\cite{wang2026odyssey}, heterogeneous collaborative navigation~\cite{wang2026can}, multi-drone perception and reasoning~\cite{zha2026aircopbench}, and LLM-based robot coordination~\cite{liu2025coherent}. 
These works highlight the importance of shared scene understanding in embodied robot teams, but they mainly operate at the level of navigation, perception, dialogue, or task planning. 
A parallel line of work studies 3DGS as a shared representation for cross-agent mapping, but its main objective is usually geometric rather than semantic consistency. 
Pairwise Gaussian registration methods align independently reconstructed Gaussian models using geometry-based coarse-to-fine registration~\cite{chang2024gaussreg}, scale-aware photometric refinement~\cite{yuan2024photoreg}, or distributional Gaussian alignment combined with photometric and depth constraints~\cite{cheng2025reggs}.
Multi-agent Gaussian SLAM systems extend single-agent Gaussian mapping to collaborative reconstruction, using intra-agent and inter-agent Gaussian consensus~\cite{xu2025macego}, local submap optimization with loop closure~\cite{thomas2025grandslam}, or Gaussian map merging for global consistency~\cite{yugay2025magicslam}.
More recent semantic multi-robot Gaussian systems incorporate language-aware or semantic cues for metric-semantic mapping and initialization-free registration~\cite{yu2025hammer,shorinwa2025siren}. 
Beyond multi-agent mapping, related Gaussian methods address sparse-view reconstruction with large-model priors~\cite{yu2024lm_gaussian}, sparse-view language Gaussian splatting~\cite{chen2025slgaussian}, feed-forward language Gaussian reconstruction from sparse unposed images~\cite{liu2026langflash}, and embodied exploration and reasoning~\cite{yu2026gaussexplorer}.
Together, these works demonstrate that Gaussian maps can be registered, fused, and queried across robotic settings. 
However, they do not explicitly target the post-fusion referability problem studied here: whether independently constructed semantic Gaussian maps remain cross-agent instance-consistent, natural-language queryable, and suitable for view-conditioned compositional and relational grounding after fusion.

\section{Methodology}
\subsection{Overview}
As shown in Fig.~\ref{fig:Pipeline}, we consider two agents
$a\in\{A,B\}$ that independently observe complementary regions of
the same indoor scene and reconstruct local semantic Gaussian maps
$\mathcal{G}^A$ and $\mathcal{G}^B$.
At inference, each query consists of a single agent observation
$(I^v,\pi^v)$ and a single referring expression $q$. Here, $I^{v}$ denotes the RGB image observed by the querying agent, and $\pi^{v}$ is the camera viewpoint. The query is
applied  to either the corresponding local map or the fused
cooperative map. Each query is applied to either a local map or the aligned cooperative maps.
Map alignment is query-independent, while cross-agent candidate association is performed at query time; the querying pose is transformed to the common frame for cooperative grounding.

CoRef-GS consists of three components.
(i) \textbf{Instance-aware Local Semantic Mapping} independently
constructs Gaussian maps with CLIP-aligned semantics and instance
grouping.
(ii) \textbf{View-conditioned Relation Reasoning} parses the
referring expression and grounds it over instance masks rendered from
the querying viewpoint.
(iii) \textbf{Cross-Agent Alignment and Instance Fusion} aligns the
local maps through image-assisted coarse-to-fine $\mathrm{Sim}(3)$
registration and associates duplicate cross-agent candidates on the
aligned maps based on geometric and semantic consistency.
The same grounding module operates on either a local or aligned
cooperative map, without joint map reconstruction or semantic retraining.

\subsection{Instance-aware Local Semantic Map}
\label{sec:gaussian_recon}
For each agent $a$, we take its RGB image sequence $\mathcal{I}^{a}$ as input and independently reconstruct an instance-aware semantic Gaussian map $\mathcal{G}^{a}$ following OpenGaussian~\cite{wu2024opengaussian}, where camera poses are estimated as part of the reconstruction pipeline.
Each Gaussian primitive is represented as
\begin{equation}
g_i^a=(\theta_i^a,\psi_i^a), \qquad 
\psi_i^a=(\mathbf{f}_i^a,z_i^a,\mathbf{e}_{z_i^a}^a),
\end{equation}
where $\theta_i^a$ denotes the standard Gaussian parameters\cite{kerbl20233d}.
Here, $\mathbf{f}_i^a$ denotes the instance feature, $z_i^a$ is the codebook assignment, and $\mathbf{e}_{z_i^a}^a$ is the CLIP-aligned semantic embedding.
Non-empty codebook groups define local leaf instances $\mathbb{I}^{a}$, which support open-vocabulary grounding and relation reasoning on both local and aligned maps.
Since semantic embeddings from all agents are anchored to the same CLIP space, independently reconstructed maps remain semantically comparable without cross-agent joint training or post-alignment feature adaptation. This shared semantic space is essential for preserving open-vocabulary queryability after map alignment and fusion.

\subsection{View-conditioned Relation Reasoning}
\label{sec:TRO}
Given a selected semantic Gaussian map $\mathcal{G}^{m}$ and its
leaf-instance set $\mathcal{I}^{m}$, derived from the instance-aware
local semantic maps constructed in Sec.~\ref{sec:gaussian_recon}, where
$m\in\{A,B,F\}$ denotes Agent A, Agent B, or their fused cooperative
map, we ground a referring expression $q$ from a querying observation
$(I^{v},\pi^{v})$.
We first retrieve semantically plausible target and landmark candidates,
and then resolve their spatial relation over instance masks rendered from
the querying viewpoint $\pi^{v}$. The same grounding procedure is applied
to both local and fused maps.

\noindent\textbf{Query Parsing and Candidate Retrieval.}
We parse $q$ into a spatial triplet $(T,R,O)$ using a lightweight
DistilBERT-based~\cite{sanh2019distilbert} spatial-role parser trained on
SpRL-2012~\cite{kordjamshidi2012semeval}, where $T$, $R$, and $O$
denote the target, relation, and landmark, respectively.
For each entity phrase $t\in\{T,O\}$, we evaluate CLIP text
similarity against the semantics of leaf instances in
$\mathcal{I}^{m}$, relative to the canonical negative set
$\{\text{object},\text{things},\text{stuff},\text{texture}\}$
following LangSplat~\cite{qin2024langsplat}.
High-confidence leaf instances are then aggregated into object-level
candidates according to instance-feature affinity.
For the fused map, candidate groups sharing the same cross-map global
identity are merged into a single candidate, yielding
$\mathcal{C}_{T}$ and $\mathcal{C}_{O}$.

\noindent\textbf{View-conditioned Mask Relation Graph (VMRG).}
For each candidate
$C_i\in\mathcal{C}_{T}\cup\mathcal{C}_{O}$,
we render its instance-specific Gaussian subset
$\mathcal{G}_{C_i}^{m}$ from the querying viewpoint
$\pi^v$ to obtain a pixel-wise opacity map $\alpha_i^v$
and depth map $D_i^v$:
\begin{equation}
(\alpha_i^v,D_i^v)
=
\mathcal{R}_{\alpha,D}(\mathcal{G}_{C_i}^{m},\pi^v),
\quad
M_i^v=\mathbf{1}[\alpha_i^v>\tau_\alpha].
\label{eq:mask_render}
\end{equation}
Here, $M_i^v$ is the binary mask of candidate $C_i$ in the querying
view, obtained by thresholding the rendered opacity map with
$\tau_\alpha$, while $D_i^v$ is the per-pixel rendered depth map of
candidate $C_i$ measured from the querying viewpoint $\pi^v$.
Candidates with non-empty masks form the visible target and landmark sets
$\mathcal{C}_{T}^{v}$ and $\mathcal{C}_{O}^{v}$, respectively.
We organize them into a directed target--landmark relation graph
$\mathcal{H}^{v}=(\mathcal{V}^{v},\mathcal{E}^{v})$, where
\begin{equation}
\mathcal{V}^{v}
=
\mathcal{C}_{T}^{v}\cup\mathcal{C}_{O}^{v},
\qquad
\mathcal{E}^{v}
=
\{(i,j)\mid
i\in\mathcal{C}_{T}^{v},
j\in\mathcal{C}_{O}^{v}\}.
\label{eq:relation_graph}
\end{equation}
Thus, each directed edge represents a candidate spatial relation from
a target instance to a landmark instance.

Each node combines the CLIP embedding of its masked RGB crop with
\begin{equation}
\boldsymbol{g}_{i}^{v}
=
[c_i^x,c_i^y,\bar d_i,
x_i^{\min},y_i^{\min},x_i^{\max},y_i^{\max}].
\end{equation}
which encodes its normalized mask centroid, mean depth, and bounding box.
The concatenated feature is projected to
$\boldsymbol{h}_{i}^{v}$ by a lightweight MLP.
To encode the relative spatial geometry between target and landmark
candidates, each directed edge $(i,j)\in\mathcal{E}^{v}$ is assigned
a relation feature $\boldsymbol{r}_{ij}^{v}$:
\begin{equation}
\begin{aligned}
\boldsymbol{r}_{ij}^{v}=[
&\Delta x_{ij},\Delta y_{ij},d_{ij},\Delta\bar d_{ij},
\operatorname{IoU}_{ij}, \\
&\eta_{i\rightarrow j},\eta_{j\rightarrow i},
\sin\theta_{ij},\cos\theta_{ij},
\rho_{ij}^{\mathrm{mask}},\rho_{ij}^{\mathrm{bbox}}
].
\end{aligned}
\label{eq:edge_geometry}
\end{equation}
Here, $\eta$ denotes directional mask coverage and $\rho$ relative
mask/bounding-box area; the remaining terms encode relative centroid
position, depth, overlap, and image-plane direction.
Because these features are computed after rendering from $\pi^v$,
view-dependent relations are represented in the querying robot's view
rather than a fixed global frame.

\noindent\textbf{Relation Reasoning and Offline Supervision.}
Following ReLaGS~\cite{xie2026relags}, we use an edge-aware GNN to
jointly reason over candidate appearance and relative geometry.
For relational queries, each directed target--landmark edge
$(i,j)\in\mathcal{E}^{v}$ is evaluated against the queried relation.
Let
$\boldsymbol{\phi}_R=
\operatorname{Norm}(\Phi_{\mathrm{text}}(R))$.
Each edge is encoded and ranked by
\begin{equation}
\begin{gathered}
\boldsymbol{z}_{ij}
=
\operatorname{Norm}\!\left(
\operatorname{GNN}_{\theta}
(\boldsymbol{h}_i^v,\boldsymbol{h}_j^v,\boldsymbol{r}_{ij}^v)
\right),
\\
(i^*,j^*)
=
\operatorname*{arg\,max}_{(i,j)\in\mathcal{E}^{v}}
\boldsymbol{z}_{ij}^{\top}\boldsymbol{\phi}_R .
\end{gathered}
\label{eq:relation_grounding}
\end{equation}
If the highest score is below $\tau_R$, the match is rejected; otherwise,
$M_i^*$ is returned. For non-relational queries, we return the top-scoring
visible target by open-vocabulary relevance.

The relation reasoner is trained offline on scene-disjoint Replica
RGB-D scenes using $21$ spatial predicates adapted from
SGFormer~\cite{lv2024sgformer}, including viewpoint-dependent relations
such as \emph{left of} and \emph{right of}.
Qwen3-VL~\cite{bai2025qwen3vl} assigns one or more predicates to each
visible directed instance pair from RGB, depth, and masks; it is used
only for offline supervision and is not invoked at inference.
The relation embeddings are aligned with the corresponding CLIP
predicate embeddings using a multi-positive contrastive objective.

\subsection{Cross Agent Alignment (CAA)}
\label{sec:registration}
Since the two local Gaussian maps are reconstructed~\cite{adorjan2016opensfm} independently, 
$\mathcal{G}^{A}$ and $\mathcal{G}^{B}$ lie in different local similarity frames. For cooperative referring, CAA aligns the source map $\mathcal{G}^{B}$ to the target frame of $\mathcal{G}^{A}$ by estimating a cross-agent Sim(3)
transformation
\begin{equation}
\begin{gathered}
\mathcal{T}_{B\rightarrow A}
=
(s_{BA},\mathbf{R}_{BA},\mathbf{t}_{BA})
\in \mathrm{Sim}(3),
\\
\mathcal{T}_{B\rightarrow A}(\mathbf{x}^{B})
=
s_{BA}\mathbf{R}_{BA}\mathbf{x}^{B}
+\mathbf{t}_{BA}.
\end{gathered}
\end{equation}
where $s_{BA}\in\mathbb{R}_{+}$, $\mathbf{R}_{BA}\in SO(3)$, and $\mathbf{t}_{BA}\in\mathbb{R}^{3}$. We estimate it with a coarse-to-fine registration pipeline.

\noindent\textbf{Coarse Cross-agent Registration.}
We first retrieve potentially overlapping cross-agent image pairs using
NetVLAD~\cite{arandjelovic2016netvlad}, extract SuperPoint
features~\cite{detone2018superpoint}, and establish correspondences with
LightGlue~\cite{lindenberger2023lightglue}.
The matches are geometrically verified using fundamental-matrix RANSAC.
For each verified pair $k$, we recover the relative rotation
$\hat{\mathbf{R}}_k$ and translation direction $\hat{\mathbf{d}}_k$
from the essential matrix.
Given the  local camera poses, a candidate transform $T$
induces a relative rotation $\mathbf R_k(T)$ and a normalized translation
direction $\mathbf d_k(T)$.
We obtain the coarse initialization by
\begin{equation}
\begin{aligned}
T^{0}_{B\rightarrow A}
&=
\arg\min_{T\in\mathrm{Sim}(3)}
\sum_k w_k\,\rho\Big(
\\[-1mm]
&\quad
\big\|
\operatorname{Log}(
\hat{\mathbf{R}}_k^{\top}\mathbf{R}_k(T))
\big\|_2^2
+
\lambda_t
\big\|
\mathbf{d}_k(T)-\hat{\mathbf{d}}_k
\big\|_2^2
\Big).
\end{aligned}
\label{eq:coarse_sim3}
\end{equation}
where $w_k$ weights each image pair by its pose-inlier support and
$\rho(\cdot)$ is a robust loss. Across multiple verified cross-agent pairs, the induced translation
directions depend jointly on the similarity scale and translation, allowing
the inter-map Sim(3) parameters to be estimated without metric cross-map
correspondences.

\noindent\textbf{Visibility-aware Rendering Refinement.}
The coarse estimate provides global alignment but may retain residual scale, rotation, and translation errors. Starting from $T^0_{B\rightarrow A}$, we fix both local maps and optimize only the cross-agent similarity transform.
Inspired by ~\cite{yuan2024photoreg}, for each selected target view $\pi_T^A$, we render RGB, depth, and opacity from the target map $\mathcal{G}^A$ and the transformed source map $T(\mathcal{G}^B)$. 
We denote these renderings as $(\hat{I}_A^A,\hat{D}_A^A,\hat{\alpha}_A^A)$ and $(\hat{I}_B^B(T),\hat{D}_B^B(T),\hat{\alpha}_B^B(T))$, respectively.
This enables rendering refinement without re-optimizing either local Gaussian map.

To suppress non-overlapping regions, we define the common-visibility
weight at pixel $p$ as
$w_\tau^T(p)=
\hat{\alpha}_\tau^A(p)\hat{\alpha}_\tau^B(T,p)$
and optimize
\begin{equation}
\begin{aligned}
\mathcal{L}_{\mathrm{CAA}}(T)
&=
\frac{1}{
\sum_{\tau,p} w_\tau^T(p)+\epsilon
}
\sum_{\tau,p} w_\tau^T(p)
\Big[
\\[-0.5ex]
&\quad
\lambda_{\mathrm{rgb}}
\left\|
\hat I_\tau^B(T,p)-\hat I_\tau^A(p)
\right\|_1
\\[-0.3ex]
&\quad+
\lambda_d
\left|
\hat D_\tau^B(T,p)-\hat D_\tau^A(p)
\right|
\Big].
\end{aligned}
\label{eq:caa_refine}
\end{equation}
Optimizing Eq.~\eqref{eq:caa_refine} yields the refined transformation
$T^{*}_{B\rightarrow A}$.
This stage establishes a common geometric frame but does not by itself resolve duplicate cross-map instance identities.

\noindent\textbf{Cross-map Instance Association.}
\label{para:cross_map_instance_association}
After geometric alignment, the same physical object may still appear as two independent local instances, which can introduce duplicate candidates during retrieval and relation reasoning.
For each local instance, we summarize its Gaussian support by a semantic descriptor $e_i$, a 3D centroid $c_i$, and an axis-aligned bounding box $B_i$.
For an instance $i\in\mathcal{I}^{A}$ and a transformed source instance $j\in T^{*}_{B\rightarrow A}(\mathcal{I}^{B})$, we compute
\begin{equation}
\scalebox{0.8}{$
\displaystyle
S_{ij}
=
\lambda_s\cos(e_i,e_j)
+
\lambda_c
\exp\!\left(
-\frac{\|c_i-c_j\|_2}{\sigma}
\right)
+
\lambda_b
\operatorname{IoU}_{3D}(B_i,B_j)
$}
\label{eq:instance_affinity}
\end{equation}
We greedily match instances one-to-one by descending $S_{ij}$,
accepting pairs with $S_{ij}\geq\tau_s$ and either centroid distance
$\leq\tau_c$ or 3D box IoU $\geq\tau_b$; matched pairs share a global
identity and unmatched instances remain independent.
The cooperative Gaussian map retains the aligned Gaussian primitives
from both agents:
\begin{equation}
\mathcal{G}^{F}
=
\mathcal{G}^{A}
\cup
T^{*}_{B\rightarrow A}(\mathcal{G}^{B}).
\label{eq:fused_map}
\end{equation}
Importantly, instance fusion does not average or relearn semantic
embeddings; it only associates matched local instances through their
global identities.
For cooperative referring, the grounding module operates on
$(\mathcal{G}^{F},\mathcal{I}^{F})$.
For single-agent referring, it directly operates on
$(\mathcal{G}^{A},\mathcal{I}^{A})$ or
$(\mathcal{G}^{B},\mathcal{I}^{B})$.
If a cooperative query originates from Agent~B, its querying camera pose
is transformed to the common frame using
$T^{*}_{B\rightarrow A}$ before view-conditioned rendering.

\section{Experiments}
\subsection{CoQuad-Ref Dataset.} 
\label{sec:dataset}
We evaluate our method on \textbf{CoQuad-Ref}, a dual-quadruped benchmark for cooperative referring scene understanding. The benchmark contains $15$ indoor scenes, including $8$ real-world scenes and $7$ simulated scenes. 
For each scene, two quadruped agents traverse complementary regions of the same environment, producing paired trajectories with partial cross-agent overlap and substantial viewpoint changes.
This setting enables controlled evaluation of cross-agent Gaussian-map registration, instance fusion, and viewpoint-conditioned referring segmentation.
\begin{figure}[!t]
    \centering

    \begin{subfigure}{0.98\columnwidth}
        \centering
        \includegraphics[width=\linewidth]{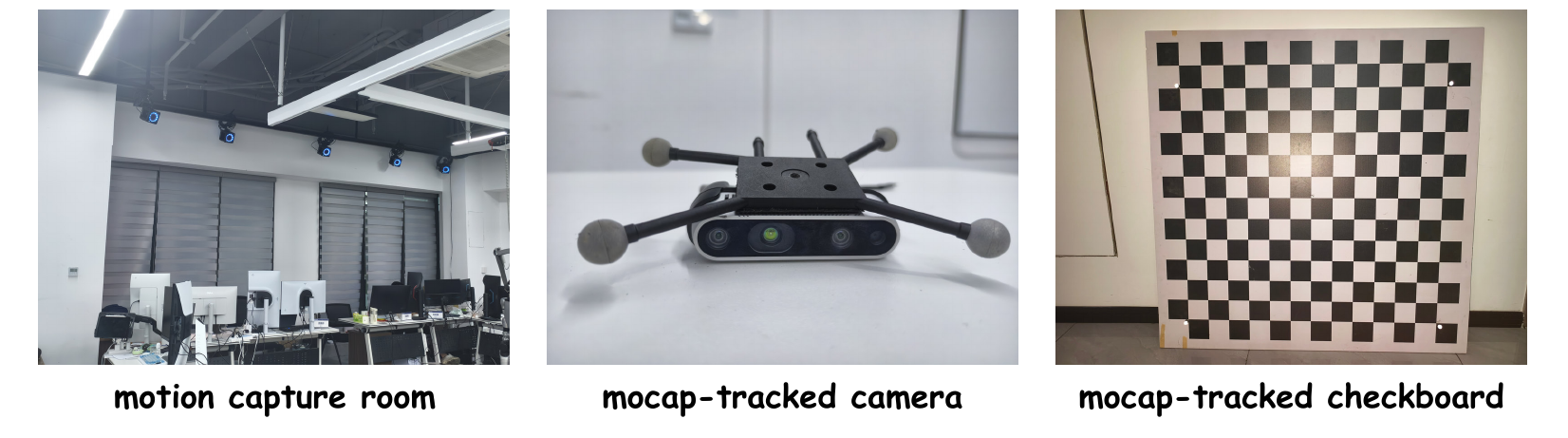}
        \vskip-1.5ex
        \caption{\textbf{Real-world data-collection setup.}}
        \vspace{-1pt}
        \label{fig:data_collection_setup}
    \end{subfigure}

    \vspace{0.8ex}

    \begin{subfigure}{0.98\columnwidth}
        \centering
        \includegraphics[width=\linewidth]{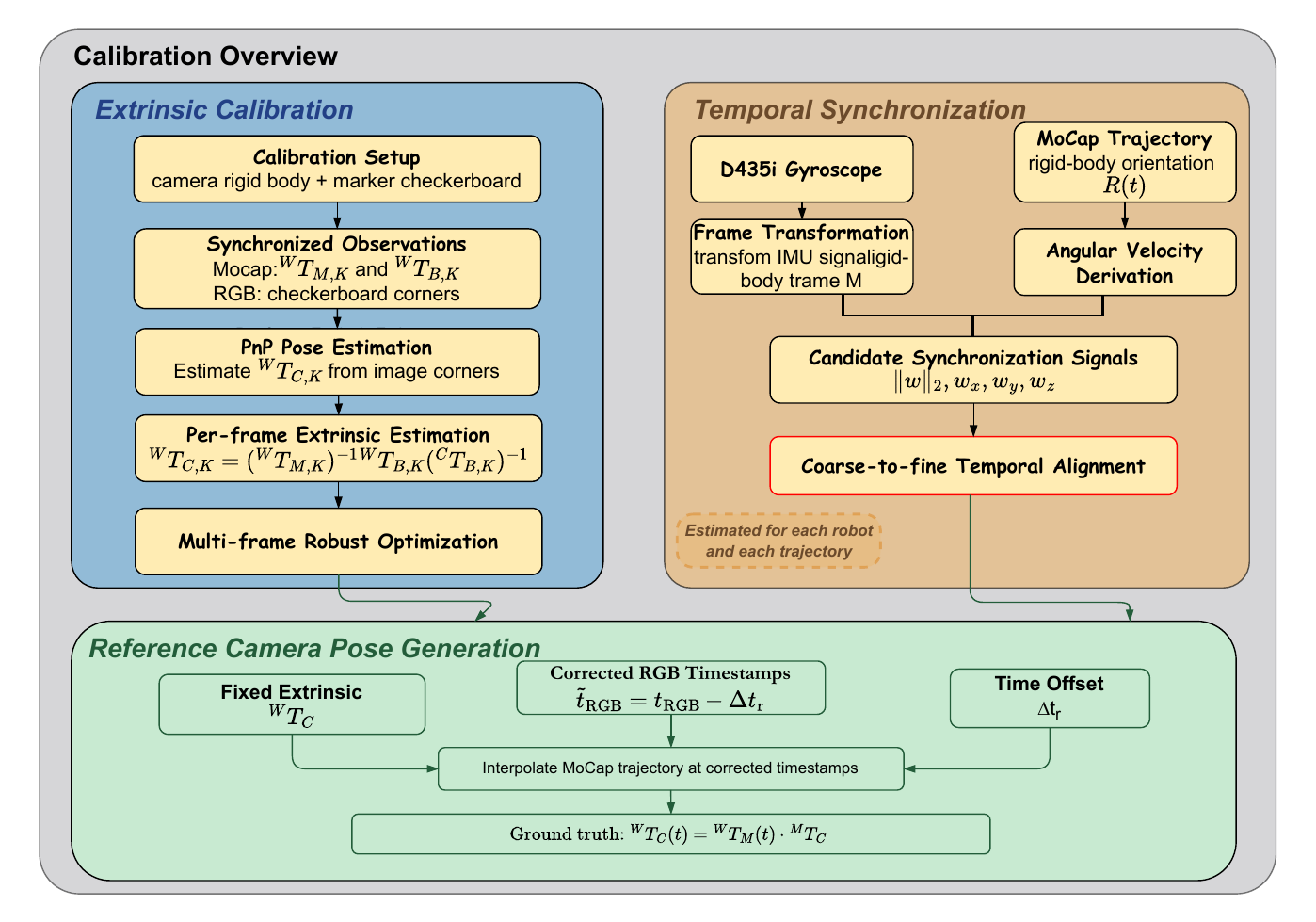}
        \vskip-1.5ex
        \caption{\textbf{Reference-pose generation pipeline.}}
        \vspace{-1pt}
        \label{fig:calibration_sync_pipeline}
    \end{subfigure}

    \caption{\textbf{Real-world data acquisition and reference-pose generation.}}
    \label{fig:realworld_pipeline}
\end{figure}

The real-world split is collected using a Unitree Go2 and a Unitree Go2-W, each equipped with an Intel RealSense D435i camera. RGB images are captured at a resolution of $960\times540$ pixels and a frame rate of $30\mathrm{Hz}$. An external motion-capture system tracks rigid marker clusters attached to the cameras at $100\mathrm{Hz}$ to provide reference poses for registration evaluation. 
The real-world data-collection platform is shown in~\cref{fig:data_collection_setup}, and the data-collection setup is shown in~\cref{fig:calibration_sync_pipeline}.
The simulated split contains one Replica and six HM3D scenes collected in Habitat-Sim using the same paired-trajectory protocol, with reference poses obtained directly from the simulator.

\noindent\textbf{Real-world Calibration and Synchronization.}
Reference camera poses are obtained from an external MoCap system after spatial and temporal calibration. We estimate the camera-MoCap extrinsic using a marker-equipped checkerboard by combining PnP camera-to-board poses with the corresponding MoCap poses and robustly refining the transformation. The resulting translation/rotation residuals are $12.84$mm/$0.904^\circ$ for Go2 and $9.37$mm/$0.821^\circ$ for Go2-W. Temporal offsets between the D435i and MoCap clocks are estimated by cross-correlating their angular-velocity trajectories, after which MoCap poses are interpolated at the corrected RGB timestamps. A shared visual trigger verifies a residual synchronization error below $33$ms.

\begin{figure*}[!t]
    \centering
    \includegraphics[width=\textwidth]{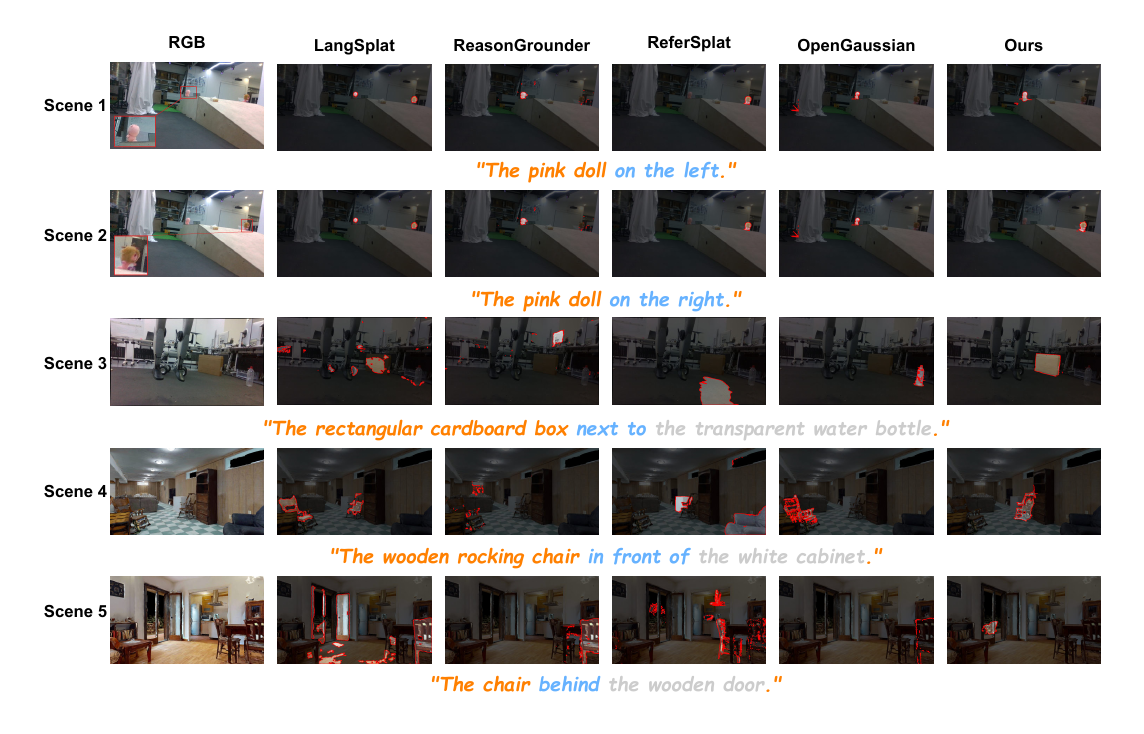}
    \vskip-4ex
    \caption{Qualitative comparison of referring segmentation results.
    The first three rows show real-world scenes, while the last two rows show simulated scenes.}
    \label{fig:openseg_compare}
\vspace{-2em}
\end{figure*}

\noindent\textbf{Evaluation Metrics.}
Since reconstructed 3DGS maps may differ from ground truth by a similarity transform, we report scale-robust rotation error, translation-direction error, and aligned ATE RMSE. 
For language grounding, following OpenGaussian~\cite{wu2024opengaussian}, we report mIoU and mAcc between predicted and ground-truth target masks.

\begin{table}[!t]
\centering
\scriptsize
\setlength{\tabcolsep}{2.4pt}
\renewcommand{\arraystretch}{1.04}

\resizebox{\columnwidth}{!}{%
\begin{tabular}{l|ccc|ccc}
\toprule
\multirow{2}{*}{\textbf{Method}}
& \multicolumn{3}{c|}{\textbf{Simulation}}
& \multicolumn{3}{c}{\textbf{Real-world}} \\
\cmidrule(lr){2-4} \cmidrule(lr){5-7}
& \textbf{Rot. $\downarrow$}
& \textbf{T-dir $\downarrow$}
& \textbf{ATE $\downarrow$}
& \textbf{Rot. $\downarrow$}
& \textbf{T-dir $\downarrow$}
& \textbf{ATE $\downarrow$} \\
\midrule

PhotoReg~\cite{yuan2024photoreg}
& 40.721 & 67.874 & 2.394
& 85.734 & 97.642 & 1.609 \\

RegGS (ID)~\cite{cheng2025reggs}
& 80.832 & 71.056 & 2.377
& 74.585 & 75.783 & 1.651 \\

RegGS (CI)~\cite{cheng2025reggs}
& 2.580 & 2.718 & 0.083
& 3.702 & 9.992 & 0.114 \\

\rowcolor{gray!12}
Ours
& \textbf{0.146} & \textbf{0.176} & \textbf{0.029}
& \textbf{3.385} & \textbf{3.085} & \textbf{0.091} \\

\bottomrule
\end{tabular}%
}

\caption{
\textbf{Gaussian map registration.}
RegGS (CI): our coarse Sim(3) initialization; RegGS (ID): identity initialization.
}
\label{tab:registration_comparison}
\vspace{-3em}
\end{table}

\begin{table}[!t]
\centering
\scriptsize
\setlength{\tabcolsep}{2.2pt}
\renewcommand{\arraystretch}{1.08}

%
\begin{minipage}[t]{\columnwidth}
\centering
\subcaption{Experimental results of simulated scenes.}

\resizebox{\columnwidth}{!}{%
\begin{tabular}{l|c|ccccccc|c}
\toprule
\textbf{Method} & \textbf{Metric}
& \textbf{S1} & \textbf{S2} & \textbf{S3} & \textbf{S4}
& \textbf{S5} & \textbf{S6} & \textbf{S7}
& \textbf{Mean} \\
\midrule

OpenGaussian~\cite{wu2024opengaussian}
& \multirow{8}{*}{mIoU$\uparrow$}
& 21.6 & 20.6 & 10.6 & 6.0 & 12.4 & 21.4 & 3.2 & 13.7 \\

LangSplat~\cite{qin2024langsplat}
&
& 29.4 & 19.8 & 40.6 & 30.1 & 15.7 & 29.0 & 13.3 & 25.4 \\

ReasonGrounder~\cite{liu2025reasongrounder}
&
& 15.8 & 19.7 & 24.3 & 52.2 & 24.8 & 31.7 & 11.0 & 25.6 \\

ReferSplat~\cite{he2025refersplat}
&
& 34.1 & 29.3 & 31.2 & 34.9 & 45.6 & 27.5 & 46.4 & 35.6 \\

ReLaGS~\cite{xie2026relags}
&
& 18.96 & 19.41 & 17.81 & 16.62 & 23.00 & 19.09 & 15.04 & 18.56 \\

LangSplatV2~\cite{li2025langsplatv2}
&
& 32.6 & 26.1 & 49.0 & 48.0 & 32.1 & 45.0 & 32.6 & 37.9 \\

Ours(SigLIP2~\cite{tschannen2025siglip})
&
& 40.7 & 39.2 & 35.3 & 32.1 & 39.1 & 35.0 & 33.4 & 36.4 \\

\rowcolor{gray!12}
Ours
&
& 49.4 & 39.9 & 50.0 & 34.3 & 59.8 & 45.8 & 43.1 & 46.0 \\

\midrule

OpenGaussian~\cite{wu2024opengaussian}
& \multirow{8}{*}{mAcc$\uparrow$}
& 23.3 & 24.7 & 12.1 & 6.9 & 16.3 & 22.8 & 3.2 & 15.6 \\

LangSplat~\cite{qin2024langsplat}
&
& 59.8 & 57.6 & 58.9 & 52.9 & 65.9 & 57.3 & 41.1 & 56.2 \\

ReasonGrounder~\cite{liu2025reasongrounder}
&
& 18.5 & 21.4 & 28.6 & 52.6 & 30.4 & 34.8 & 15.8 & 28.9 \\

ReferSplat~\cite{he2025refersplat}
&
& 70.2 & 69.3 & 65.7 & 75.5 & 83.1 & 72.4 & 79.6 & 73.7 \\

ReLaGS~\cite{xie2026relags}
&
& 41.28 & 49.80 & 23.14 & 35.89 & 39.45 & 44.71 & 28.66 & 37.56 \\

LangSplatV2~\cite{li2025langsplatv2}
&
& 75.4 & 73.9 & 78.7 & 82.5 & 71.5 & 79.1 & 63.4 & 74.9 \\

Ours (SigLIP2~\cite{tschannen2025siglip})
&
& 62.9 & 45.6 & 35.0 & 46.8 & 69.2 & 56.2 & 41.0 & 51.0 \\

\rowcolor{gray!12}
Ours
&
& 73.2 & 62.1 & 69.0 & 48.2 & 77.2 & 68.7 & 50.1 & 64.1 \\

\bottomrule
\end{tabular}%
}
\end{minipage}

\vspace{1em}

%
\begin{minipage}[t]{\columnwidth}
\centering
\subcaption{Experimental results of real-world scenes.}

\resizebox{\columnwidth}{!}{%
\begin{tabular}{l|c|cccccccc|c}
\toprule
\textbf{Method} & \textbf{Metric}
& \textbf{S1} & \textbf{S2} & \textbf{S3} & \textbf{S4}
& \textbf{S5} & \textbf{S6} & \textbf{S7} & \textbf{S8}
& \textbf{Mean} \\
\midrule

OpenGaussian~\cite{wu2024opengaussian}
& \multirow{8}{*}{mIoU$\uparrow$}
& 36.6 & 29.1 & 63.0 & 34.2 & 8.7 & 23.7 & 40.8 & 13.0
& 31.1 \\

LangSplat~\cite{qin2024langsplat}
&
& 20.8 & 30.4 & 11.0 & 4.4 & 0.8 & 0.4 & 10.3 & 3.8
& 10.2 \\

ReasonGrounder~\cite{liu2025reasongrounder}
&
& 8.3 & 0.2 & 0.0 & 19.8 & 25.6 & 8.7 & 0.1 & 28.8
& 11.4 \\

ReferSplat~\cite{he2025refersplat}
&
& 35.0 & 79.5 & 75.9 & 60.0 & 46.5 & 39.4 & 33.3 & 51.1
& 52.6 \\

ReLaGS~\cite{xie2026relags}
&
& 22.04 & 21.22 & 33.47 & 27.76 & 47.04 & 12.17 & 18.94 & 21.55
& 25.52 \\

LangSplatV2~\cite{li2025langsplatv2}
&
& 32.8 & 36.5 & 56.1 & 30.2 & 22.7 & 26.2 & 35.9 & 53.0
& 36.7 \\

Ours (SigLIP2~\cite{tschannen2025siglip})
&
& 47.3 & 59.4 & 63.1 & 65.7 & 58.5 & 43.1 & 69.4 & 56.6
& 57.9 \\

\rowcolor{gray!12}
Ours
&
& 70.1 & 73.3 & 76.7 & 62.4 & 67.4 & 61.1 & 79.8 & 59.9
& 68.8 \\

\midrule

OpenGaussian~\cite{wu2024opengaussian}
& \multirow{8}{*}{mAcc$\uparrow$}
& 41.6 & 29.6 & 71.5 & 37.3 & 9.7 & 27.2 & 46.0 & 20.3
& 35.4 \\

LangSplat~\cite{qin2024langsplat}
&
& 36.1 & 33.9 & 49.6 & 21.5 & 10.8 & 9.7 & 38.2 & 39.3
& 29.9 \\

ReasonGrounder~\cite{liu2025reasongrounder}
&
& 20.6 & 1.4 & 0.0 & 21.9 & 37.6 & 18.0 & 2.0 & 45.3
& 18.3 \\

ReferSplat~\cite{he2025refersplat}
&
& 68.5 & 96.1 & 96.4 & 82.0 & 74.6 & 82.5 & 72.9 & 86.5
& 82.4 \\

ReLaGS~\cite{xie2026relags}
&
& 48.18 & 69.39 & 78.12 & 73.86 & 86.92 & 26.14 & 62.78 & 58.20
& 62.95 \\

LangSplatV2~\cite{li2025langsplatv2}
&
& 80.1 & 57.8 & 91.9 & 56.5 & 72.4 & 63.9 & 67.2 & 77.0
& 70.9 \\

Ours (SigLIP2~\cite{tschannen2025siglip})
&
& 65.2 & 99.0 & 76.5 & 84.4 & 79.4 & 75.2 & 93.5 & 75.7
& 81.1 \\

\rowcolor{gray!12}
Ours
&
& 87.3 & 91.4 & 97.0 & 90.7 & 91.0 & 85.1 & 92.0 & 77.1
& 89.0 \\

\bottomrule
\end{tabular}%
}
\end{minipage}

\caption{
Quantitative evaluation of instance-level spatial perception on our CoQuad-Ref dataset.
(a) Simulated scenes with seven scenes.
(b) Real-world scenes with eight scenes.
}
\label{tab:openseg_comparison}

\end{table}

\noindent\textbf{Implementation Details.}
All experiments are conducted on a single NVIDIA RTX 4090 GPU.
Local mapping follows OpenGaussian~[13]. The parser is trained for 60 epochs (batch size 8, LR $3\times10^{-5}$), and the relation reasoner uses a scene-disjoint 13/4 Replica split ($\tau=0.07$, 32 negatives/edge). For CAA, we retrieve 10 NetVLAD pairs and retain SuperPoint–LightGlue pairs with $\geq30$ RANSAC inliers and $\geq0.25$ inlier ratio. Sim(3) refinement uses Adam for 500 iterations (LR $10^{-2}$, $\lambda_{\rm rgb}=1$, $\lambda_d=0.2$). We use $\tau_{\rm rel}=0.4$, $\tau_{\rm inst}=0.9$,
$\tau_\alpha=0.4$, and $\tau_R=0.05$.
For cross-agent candidate association, we set
$(\lambda_s,\lambda_c,\lambda_b)=(0.45,0.30,0.15)$,
$\sigma=0.5$, $\tau_s=0.45$, $\tau_c=0.75$, and
$\tau_b=0.01$.

\begin{table}[!t]
\centering
\scriptsize
\setlength{\tabcolsep}{1.8pt}
\renewcommand{\arraystretch}{1.05}
\vspace{-2em}
%
\begin{minipage}[t]{\linewidth}
\centering
\subcaption{Map-alignment compatibility.}
\label{tab:alignment_open_vocab_comparison}
\vspace{0.2ex}

\resizebox{\linewidth}{!}{%
\begin{tabular}{lccccc}
\toprule
\multirow{2}{*}{\textbf{Method}}
& \multirow{2}{*}{$B$}
& \multicolumn{2}{c}{$B{\rightarrow}A$}
& \multicolumn{2}{c}{\textbf{Change}} \\
\cmidrule(lr){3-4}
\cmidrule(lr){5-6}
&
& \textbf{A net}
& \makecell{\textbf{$B$ net}\\\textbf{(coordinate only)}}
& $\Delta_{\mathrm{full}}$
& $\Delta_{\mathrm{coord.}}$ \\
\midrule

ReferSplat~\cite{he2025refersplat}
& 38.99
& 18.88
& 38.19
& -51.58\%
& -2.05\% \\

\rowcolor{gray!12}
\textbf{Ours}
& \textbf{40.96}
& \textbf{40.96}
& \textbf{40.96}
& \textbf{0.00\%}
& \textbf{0.00\%} \\

\bottomrule
\end{tabular}%
}

\end{minipage}

\vspace{0.8ex}

%
\begin{minipage}[t]{\linewidth}
\centering
\subcaption{Local maps \textit{vs.} cooperative aligned map.}
\label{tab:ablation_openseg_cooperate}
\vspace{0.2ex}

\resizebox{\linewidth}{!}{%
\begin{tabular}{lccccc}
\toprule
\multirow{2}{*}{\textbf{Setting}}
& \multicolumn{2}{c}{\textbf{Map Source}}
& \multicolumn{2}{c}{\textbf{Performance}}
& \multirow{2}{*}{\textbf{Gain}} \\
\cmidrule(lr){2-3}
\cmidrule(lr){4-5}
& \textbf{Agent A}
& \textbf{Agent B}
& \textbf{mIoU$\uparrow$}
& \textbf{mAcc$\uparrow$}
& \\
\midrule

Agent A Local Map
& \checkmark & --
& 41.91\%
& 58.58\%
& -- \\

Agent B Local Map
& -- & \checkmark
& 36.37\%
& 49.26\%
& -- \\

\rowcolor{gray!12}
Cooperative Aligned Map
& \checkmark & \checkmark
& \textbf{47.54\%}
& \textbf{74.73\%}
& \textbf{+13.4\% / +27.6\%} \\

\bottomrule
\end{tabular}%
}

\end{minipage}

\vskip-1ex
\caption{
Analysis of map-alignment compatibility and cooperative view complementarity.
(a) Open-vocabulary segmentation under different map-alignment settings
(mIoU, \% $\uparrow$).
(b) Comparison between individual local maps and the cooperative aligned map.
}
\label{tab:alignment_cooperation_analysis}
\vspace{-2em}
\end{table}

\begin{figure*}[!b]
    \vskip-1.5em
    \centering
    \includegraphics[width=\textwidth]{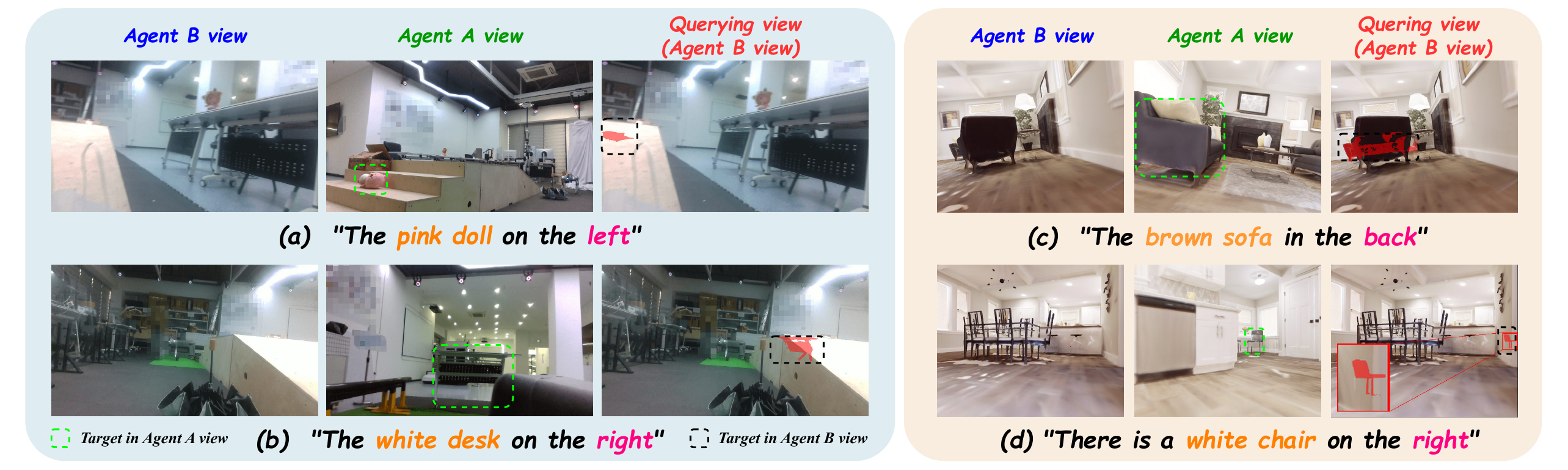}
    \vspace{-1.5em}
    \caption{Visualization of cooperative multi-agent referring segmentation.}
    \label{fig:cooperative_ref}
\end{figure*}

\subsection{Main Results and Ablations}
\noindent\textbf{Results of 3D Gaussian Registration.}
As shown in~\cref{tab:registration_comparison}, CoRef-GS achieves the lowest registration errors on both simulated and real-world scenes.
Despite large viewpoint gaps and limited cross-agent overlap, our coarse $\mathrm{Sim}(3)$ initialization substantially improves RegGS, while the full method further reduces the simulated errors to $0.146^\circ$ rotation, $0.176^\circ$ translation direction, and $0.029$ ATE RMSE.

\noindent\textbf{Results of Open-vocabulary Querying.}
CoRef-GS achieves the best referring-segmentation performance on both real and simulated scenes (\cref{tab:openseg_comparison,fig:openseg_compare}), demonstrating the benefit of reasoning over view-conditioned target-landmark relations rather than semantic similarity alone.
Replacing the image encoder with SigLIP2~\cite{tschannen2025siglip} requires no architectural changes; its stronger background responses introduce more false positives and slightly reduce accuracy, while confirming that CoRef-GS is not tied to a specific VLM backbone.

We further present cooperative multi-agent referring segmentation results in ~\cref{fig:cooperative_ref}, where aligned observations enable agents to ground queries beyond the coverage of a single robot.
As analyzed in~\cref{tab:ablation_openseg_cooperate}, cooperative fusion improves referring segmentation by integrating complementary observations from different agents.
Finally, we evaluate whether open-vocabulary queryability is preserved when independently reconstructed maps are brought into a common frame. 
As shown in~\cref{tab:alignment_open_vocab_comparison}, ReferSplat~\cite{he2025refersplat} decreases from $38.99\%$ to $18.88\%$ mIoU when the aligned map is queried under the other map's referring space, corresponding to a $51.58\%$ relative drop, whereas CoRef-GS preserves $40.96\%$ mIoU without retraining. 
To isolate the effect of coordinate transformation itself, we additionally transform ReferSplat's map and query cameras while retaining its original querying network. 
Its mIoU remains at $38.19\%$, only $2.05\%$ below the original result. 
This control confirms that the large cross-map degradation is not caused by the coordinate transformation itself, but primarily by incompatibility between independently learned semantic query spaces.
In contrast, our shared CLIP-aligned representation remains directly comparable across independently reconstructed maps.
\begin{table}[!t]
\centering
\scriptsize
\setlength{\tabcolsep}{2.0pt}
\renewcommand{\arraystretch}{1.05}
%
\begin{minipage}[t]{0.52\linewidth}
\centering
\subcaption{Photometric optimization.}
\label{tab:po_ablation_sim}
\vspace{0.2ex}

\resizebox{0.9\linewidth}{!}{%
\begin{tabular}{lccc}
\toprule
\textbf{Method}
& \textbf{Rot. ($^\circ$) $\downarrow$}
& \textbf{Trans. ($^\circ$) $\downarrow$}
& \textbf{ATE RMSE $\downarrow$} \\
\midrule

w/o PO
& 0.205
& 0.194
& 0.056 \\

\rowcolor{gray!12}
\textbf{w/ PO}
& \textbf{0.146}
& \textbf{0.176}
& \textbf{0.050} \\

\bottomrule
\end{tabular}%
}

\end{minipage}%
\hfill%
%
\begin{minipage}[t]{0.46\linewidth}
\centering
\subcaption{Instance fusion.}
\label{tab:instance_fusion_ablation}
\vspace{0.2ex}

\resizebox{\linewidth}{!}{%
\begin{tabular}{lccc}
\toprule
\textbf{Variant}
& \textbf{mIoU (\%)}
& \textbf{mAcc (\%)}
& \textbf{Gain (pp)} \\
\midrule

w/o Fusion
& 39.06
& 70.09
& -- \\

\rowcolor{gray!12}
\textbf{Full}
& \textbf{47.54}
& \textbf{74.73}
& \textbf{+8.48/+4.64} \\

\bottomrule
\end{tabular}%
}

\end{minipage}

\vspace{-0.5ex}
\caption{\textbf{Ablation studies on key components.}
(a) Effect of photometric optimization on map alignment on the simulation split.
(b) Effect of instance fusion on cooperative referring segmentation.}
\label{tab:component_ablation}
\vspace{-2.5em}
\end{table}

\noindent\textbf{Ablation Studies.}
Photometric refinement consistently improves coarse Sim(3) estimate on simulated scenes~(\cref{tab:po_ablation_sim}), reducing residual rotation, translation-direction, and trajectory errors after correspondence-based initialization. This indicates that sparse feature correspondences provide reliable global initialization, while visibility-aware RGB-depth consistency enables fine alignment.

For cooperative referring, the fused map reaches $47.54\%$ mIoU and $74.73\%$ mAcc, outperforming either local map~(\cref{tab:ablation_openseg_cooperate}).
Removing cross-map instance association while retaining registration, CLIP semantics, and relation reasoning reduces mIoU from $47.54\%$ to $39.06\%$ (\cref{tab:instance_fusion_ablation}). 
This confirms that geometric alignment alone is insufficient: duplicate local identities produce redundant target and landmark candidates, whereas instance association enables consistent object-level relational grounding.

\section{Conclusion}
\label{sec/conclusion}
We introduced \textbf{CoRef-GS}, a cooperative referring Gaussian splatting framework for multi-robot language-grounded scene understanding.
CoRef-GS supports referring segmentation over local and aligned Gaussian maps by combining instance-aware semantic mapping, cross-agent registration, instance fusion, and view-conditioned spatial relational reasoning.
We also presented \textbf{CoQuad-Ref}, a cooperative quadruped referring benchmark covering both simulated and real-world indoor scenes.
Experiments demonstrate improved cross-agent alignment and more reliable grounding under complementary robot observations.

\bibliographystyle{IEEEtran}
\bibliography{bib}

\end{document}